\documentclass[times,review,10pt]{elsarticle}

\usepackage[UTF8]{ctex}
\usepackage{amssymb}
\usepackage{amsmath}
\usepackage{mathtools}
\usepackage{bm}
\usepackage{graphicx}
\usepackage{multirow}
\usepackage{booktabs}
\usepackage{array}
\usepackage{tabularx}
\usepackage{arydshln}
\usepackage{enumitem}
\usepackage{subfig}
\usepackage{algorithm}
\usepackage{algpseudocode}
\usepackage{hyperref}
\hypersetup{hidelinks}
\graphicspath{{./}{pic/}}
\floatname{algorithm}{Algorithm}
\algrenewcommand\algorithmicrequire{Input:}
\algrenewcommand\algorithmicensure{Output:}
\journal{Pattern Recognition}

\begin{document}

\begin{frontmatter}

\title{Interpretable Multi-Hypersphere Deep Anomaly Detection  for Open-set Supervised Anomaly Detection}
\author[aff1]{Zhiji Yang\corref{cor1}}
\ead{yangzhiji@ynufe.edu.cn}
\cortext[cor1]{Corresponding author}
\author[aff1]{Fangyong Wang}
\author[aff1]{Yue Li}
\author[aff2]{Xianli Pan}
\author[aff1]{Jianhua Zhao}
\affiliation[aff1]{organization={School of Statistics and Mathematics, Yunnan University of Finance and Economics},
	city={Kunming},
	postcode={650221},
	country={China}}
\affiliation[aff2]{organization={Beijing National Center for Applied Mathematics, Capital Normal University},
	city={Beijing},
	postcode={100048},
	country={China}}

\begin{abstract}
Multi-class open-set anomaly detection requires a model to characterize the normal acceptance domain formed by multiple heterogeneous subdistributions using only class-labeled samples from known normal classes, and to identify previously unseen anomalies at test time. Existing single-hypersphere methods cannot explicitly represent class-specific locations and acceptance ranges, while current multi-hypersphere or multi-class approaches do not fully integrate inter-class boundary constraints, learnable acceptance ranges, and interpretable decisions. To address these limitations, we propose Interpretable Multi-Hypersphere Deep Anomaly Detection (IMHD-AD). IMHD-AD constructs an independent hypersphere for each known normal class in a shared feature space. With target-inside and non-target-outside constraints, IMHD-AD embeds the class-specific hypersphere centers and radii directly into the final network layer and jointly optimizes them with the shared representation. The minimum signed boundary score across hyperspheres simultaneously determines open-set acceptance or rejection and provides a faithful geometric explanation of each decision. On MNIST, Fashion-MNIST, and CIFAR-10, IMHD-AD achieves the highest AUC in 28 of 30 open-set comparisons. A two-dimensional synthetic study further shows that model architecture must balance the compactness of known normal classes against the separability of unknown anomalies.
\end{abstract}

\begin{keyword}
\sep Anomaly detection \sep Open-set recognition
\sep Deep multi-hypersphere learning \sep Support vector data description
\sep Interpretable deep learning
\end{keyword}

\end{frontmatter}


\section{Introduction}

Anomaly detection identifies samples that deviate from known normal patterns. It is important in applications such as industrial visual inspection and medical image analysis \cite{batzner2024efficientad,cai2025medianomaly}. In many practical settings, training data are available for several known normal classes. Their class labels are also available, but representative labeled anomalies are difficult to collect. Normal data may also contain subdistributions with different locations, scales, and shapes. These differences can arise from object categories, operating conditions, or individual characteristics. A useful model must characterize this heterogeneous normal population without labeled anomalies. It must also preserve the structure of different normal modes and reject previously unseen classes at test time. We consider this open-set setting. The training set contains only class-labeled samples from known normal classes, and unknown anomaly classes are absent during training.

Data-description methods detect anomalies by estimating the support of normal data. Support Vector Data Description (SVDD) represents the acceptance region of a target distribution with a hypersphere \cite{tax2004support}. DeepSVDD combines this description with deep representation learning for high-dimensional nonlinear data \cite{ruff2018deep}. However, a single center and a shared radius cannot explicitly represent heterogeneous normal subdistributions. Enlarging one sphere to cover separated normal regions may also include unsupported regions between them. Excessive contraction may instead obscure their structure and exclude valid normal samples. A deep mapping can reshape the input distribution, but a single-center model still cannot separately adapt the location and acceptance range of each normal mode. The normal acceptance domain should therefore be modeled as a union of local regions.

Multi-hypersphere data description offers a natural way to model heterogeneous normal data. DMSVDD uses multiple hyperspheres to represent multimodal normal structure \cite{ghafoori2020deep}. Later work uses disentangled representations to reduce interference from irrelevant features \cite{xing2024deep}. However, adding more hyperspheres does not by itself produce discriminative boundaries aligned with known normal classes. A local compactness objective can cluster samples around assigned spheres. It may not prevent one known normal class from entering the acceptance region of another. Conventional multi-class SVDD improves class discrimination by incorporating samples from other classes \cite{mu2009multiclass}. DeepMAD treats the remaining known classes as non-target samples for the current target class. It learns discriminative representations through center attraction and distance-margin constraints \cite{singh2022multi}. These non-target samples still belong to the overall normal training set. They are not labeled anomalies. DeepMAD also does not jointly learn a class-specific radius as an explicit acceptance boundary. A gap therefore remains. A deep multi-hypersphere model is needed to combine class-level local description, inter-class boundary constraints, and learnable acceptance ranges using only class-labeled normal data.

Interpretability is also important for heterogeneous normal data. A model should indicate which known normal mode accepts a sample. It should also explain why all known modes reject an anomalous sample. Our previous work, Interpretable Maximum Margin Deep Anomaly Detection (IMD-AD) \cite{yang2026interpretablemaximummargindeep}, provides a useful starting point. IMD-AD embeds a hypersphere center and radius into the final network layer. This design couples boundary learning with feature learning and gives the model a geometric interpretation. However, IMD-AD uses a small number of labeled anomalies. It also constructs a normal inner sphere and a concentric margin boundary around a single center. It cannot assign separate centers and acceptance ranges to multiple normal subdistributions. Intrinsically interpretable models require explanations to follow directly from their computations \cite{rudin2019stop}. Interpretable neural clustering also shows that geometric prototypes can be encoded in network parameters \cite{JMLR:v23:19-497}. These observations motivate a multi-hypersphere extension of the geometric parameterization in IMD-AD. The extension should learn class-specific boundaries from class-labeled normal data alone. Its centers, radii, and boundary scores should directly determine both acceptance and rejection.

Accordingly, we propose IMHD-AD, an interpretable multi-hypersphere approach to deep anomaly detection. IMHD-AD constructs one hypersphere for each known normal class in a shared feature space. The union of their interiors forms the overall normal acceptance domain. For the $k$th hypersphere, samples from class $k$ impose inside-sphere constraints. Samples from the other known normal classes impose outside-sphere constraints. All of these samples are normal training data. The other classes provide relative boundary information rather than anomaly supervision. IMHD-AD jointly optimizes the feature network, the hypersphere radii, and both types of boundary-violation loss. It thereby learns local compactness, class separability, and class-specific acceptance ranges. The model also extends the parameter mapping of a single hypersphere to multiple output units. This design updates all centers and radii together with the shared representation. At test time, any sample inside a known-class hypersphere is accepted by the corresponding normal mode. A sample outside all hyperspheres is identified as an unknown anomaly. The decision depends only on the learned normal acceptance domain and does not require a prior model of the anomaly distribution.

\noindent\textbf{The main contributions of this work are summarized as follows:}
\begin{enumerate}[label={(\arabic*)}]
    \item \textbf{Normal-only multi-hypersphere modeling of heterogeneous distributions.} IMHD-AD assigns a separate hypersphere to each known normal class in a shared feature space. The union of these local regions represents the overall normal population. No labeled anomaly is used during training. Unknown anomalies are detected at test time when they fall outside the learned normal acceptance domain.

    \item \textbf{End-to-end collaborative learning of multi-hypersphere geometry.} We extend the geometric parameterization of IMD-AD to multiple hyperspheres. The center and radius of each hypersphere are embedded into the weights and bias of the final network layer. These parameters are optimized jointly with the deep representation. The class-specific acceptance regions can therefore adapt as the representation changes.

    \item \textbf{Bidirectional inter-class boundary constraints without labeled anomalies.} A one-vs-rest formulation penalizes target-class samples outside their corresponding hypersphere. It also penalizes samples from other normal classes inside that hypersphere. The objective uses structural differences among normal classes rather than known anomalies or a normal--anomaly margin. It improves inter-class separability while preserving intra-class compactness.

    \item \textbf{An intrinsic geometric explanation consistent with the joint decision rule.} Each class-specific boundary score is the squared distance to a center minus the corresponding squared radius. The minimum score determines normal-mode acceptance or rejection. These scores show the position of a sample relative to every known-class boundary. They support traceable analysis of its acceptance source, class ambiguity, and anomaly decision.
\end{enumerate}

We evaluate IMHD-AD on MNIST, Fashion-MNIST, and CIFAR-10 under two open-set protocols. The first protocol uses two known normal classes and eight unseen anomaly classes. The second uses three known normal classes and seven unseen anomaly classes. IMHD-AD obtains the highest AUC in 13 of 15 comparisons under the first protocol. It ranks first in all 15 comparisons under the second protocol. A two-dimensional synthetic study further shows how network complexity affects feature-space geometry and the separation of unknown anomalies. These results support the use of IMHD-AD for learning acceptance domains from heterogeneous normal data without labeled anomalies.

The remainder of this paper is organized as follows. Section 2 reviews interpretable maximum-margin deep anomaly detection and related deep multi-class anomaly-detection methods. Section 3 presents the objective, network parameterization, optimization procedure, and decision rule of IMHD-AD. Section 4 reports the experimental setup, comparative results, and visualization analysis. Section 5 concludes the paper and discusses future directions.

\section{Related work}
This section reviews two lines of research related to the proposed method. We first discuss the theoretical basis, model structure, and limitations of interpretable maximum-margin deep anomaly detection. We then review multi-hypersphere descriptions and inter-class constraints in deep multi-class anomaly detection. Together, these studies motivate our geometric parameterization, class-level hypersphere modeling, and use of non-target samples.

\subsection{Interpretable maximum-margin deep anomaly detection}

DeepSVDD maps normal samples into a compact hypersphere through a deep network and identifies a test sample as anomalous when it falls outside the sphere \cite{ruff2018deep}. Its objective, however, primarily emphasizes intra-class compactness and may produce degenerate feature representations. Moreover, the center and radius are commonly determined using heuristics such as the feature mean and a distance quantile, rather than learned jointly with the representation as network parameters. This separation limits both the accuracy of boundary estimation and the intrinsic interpretability of the model.

To address these limitations, Interpretable Maximum Margin Deep Anomaly Detection (IMD-AD) is proposed \cite{yang2026interpretablemaximummargindeep}. In addition to a large set of normal samples, IMD-AD uses a small number of labeled anomalies and constructs a normal inner sphere and an outer anomaly boundary around a common center. Normal samples are constrained within a hypersphere of radius $R$, while labeled anomalies are pushed outside a concentric hypersphere of radius $\sqrt{R^2+\rho^2}$, where $\rho$ denotes the margin between the normal region and known anomalies. By shrinking the normal hypersphere, penalizing normal samples that violate the inner boundary, penalizing anomalies that enter the outer boundary, and maximizing the margin, IMD-AD unifies normal-region description with maximum-margin learning under limited anomaly supervision.

Another central idea of IMD-AD is to embed the hypersphere geometry into the final layer of a neural network. Given a feature representation $z=\phi(x;\theta)$, the score of a sample relative to the normal hypersphere boundary can be expanded as
\begin{equation}
\begin{aligned}
\|z-C\|_2^2-R^2
&=
\|z\|_2^2-2C^\top z+\|C\|_2^2-R^2.
\end{aligned}
\end{equation}
Defining the weight and bias of the final layer as
\begin{equation}
\mathbf{w}=-2C,
\qquad
b=\|C\|_2^2-R^2,
\end{equation}
allows the boundary score to be written as
\begin{equation}
\|z-C\|_2^2-R^2
=
\mathbf{w}^{\top}z+b+\|z\|_2^2.
\end{equation}
Under the unit-center constraint $\|C\|_2^2=1$, the center and radius can be recovered from the parameters of the final layer. The feature network, center, radius, and margin can therefore be optimized jointly within a single model, while the final layer acquires an explicit hypersphere interpretation. A detection result is explained by the position of a sample relative to the center, the normal boundary, and the margin boundary, yielding an intrinsic geometric explanation that is faithful to the actual decision process.

Although IMD-AD establishes a basis for end-to-end hypersphere learning and intrinsic interpretation, it remains a single-center method that requires a small set of labeled anomalies. All normal samples are represented by one hypersphere, and normal--anomaly separation is defined around a common center and an explicit margin. When the normal population contains multiple known classes with different locations and scales, this structure cannot assign an independent center and acceptance range to each subdistribution or exploit relationships among normal classes to shape multiple local boundaries. The proposed IMHD-AD inherits the neural parameterization of hypersphere geometry from IMD-AD but not its anomaly supervision or explicit maximum-margin objective. Using only class-labeled samples from known normal classes, IMHD-AD learns a separate hypersphere for each class and defines the normal acceptance domain as their union. Unknown anomalies outside all known normal regions can then be detected at test time.

\subsection{Deep multi-class anomaly detection}

Multi-class anomaly detection treats several known classes as normal and the remaining unknown classes as anomalies. It must represent distinct normal modes and prevent one known class from being accepted by the region of another. DMSVDD provides a geometric solution by learning multiple hyperspheres and sample assignments in a shared feature space \cite{ghafoori2020deep}. Its extensions improve representation quality through multi-view learning \cite{chen2024dmvsvdd}, disentangled features \cite{xing2024deep}, or Gaussian-mixture latent priors \cite{wu2024deep}. However, these methods generally partition data into local modes without using known class boundaries. Their objectives mainly promote local compactness, and sample assignments often follow the nearest center. They do not explicitly keep known non-target classes outside a target region. Their learned regions therefore need not correspond to class-specific acceptance regions.

Another line of work uses class labels to construct data-description boundaries. Early methods train one SVDD model for each class or derive class domains from separate SVDD boundaries \cite{kang2006svdd,lee2007domain}. Multi-class SVDD further uses samples from other classes to impose negative constraints on the target hypersphere \cite{mu2009multiclass}. These methods have a clear geometric meaning, but they operate on raw features or in a fixed kernel space. They cannot jointly learn nonlinear representations and class boundaries. This limitation is important for high-dimensional inputs that require task-specific feature learning.

Deep methods incorporate inter-class relationships into representation learning. DeepMAD treats the remaining known classes as pseudo-anomalies relative to the current class. These samples still belong to the overall normal training set. DeepMAD uses center attraction and distance-margin objectives to improve class separation \cite{singh2022multi}. This strategy shows that other normal classes can provide relative negative supervision. However, DeepMAD does not jointly learn a class-specific radius as an explicit acceptance boundary. MMHAD uses marginal anomaly exposure and a margin constraint to improve discrimination near normal boundaries \cite{gao2024multi}. UniAD instead uses a unified reconstruction network with a layer-wise query decoder, neighborhood-masked attention, and feature jittering \cite{you2022unified}. Its anomaly score is based on reconstruction error rather than class-specific hypersphere boundaries.

Existing methods therefore do not fully combine class-level local description, inter-class constraints, learnable acceptance ranges, and interpretable end-to-end decisions. Unsupervised multi-hypersphere methods do not ensure that other known classes remain outside a target sphere. Methods such as DeepMAD exploit non-target samples but rely on center distances or predefined margins. In particular, they do not adapt explicit class radii with the representation and then use the same boundaries for detection and explanation. IMHD-AD addresses this gap with one hypersphere per known normal class. Target samples impose inside-sphere constraints, while other known classes impose outside-sphere constraints. The centers, radii, and shared feature network are optimized jointly. Their union defines the normal acceptance domain, and the minimum boundary score provides both the anomaly decision and its geometric explanation.
\section{Interpretable multi-hypersphere deep anomaly detection}

We propose Interpretable Multi-Hypersphere Deep Anomaly Detection (IMHD-AD). The method constructs an independent hypersphere for each known class and embeds its center and radius into the final network layer, thereby unifying feature extraction, multi-hypersphere boundary learning, and anomaly detection.

\subsection{Multi-hypersphere anomaly-detection objective}

Let the training set be $\mathcal{D}=\{(x_i,y_i)\}_{i=1}^{n}$, where $y_i\in\{1,\ldots,K\}$. For class $k$, the index sets of target and non-target samples are defined as
\begin{equation}
\mathcal{I}_{k}^{+}=\{i\mid y_i=k\},
\qquad
\mathcal{I}_{k}^{-}=\{i\mid y_i\neq k\},
\end{equation}
with $n_k^{+}=|\mathcal{I}_{k}^{+}|$ and $n_k^{-}=|\mathcal{I}_{k}^{-}|$. A feature extractor $\phi(\cdot;\theta)$ maps $x_i$ to $z_i=\phi(x_i;\theta)\in\mathbb{R}^{d}$. In the resulting feature space, class $k$ is represented by a hypersphere with center $C_k$ and radius $R_k$.

DMSVDD represents normal data with multiple hyperspheres, but its objective primarily constrains hypersphere volume and the compactness of enclosed samples \cite{ghafoori2020deep}. DeepMAD instead uses samples from other classes as pseudo-anomalies to improve inter-class separation \cite{singh2022multi}. IMHD-AD incorporates both types of constraint: target samples are encouraged to lie inside their corresponding hypersphere, whereas non-target samples are excluded from it.

The boundary score of sample $x_i$ with respect to the $k$th hypersphere is defined as
\begin{equation}
s_{ik}
=
\|\phi(x_i;\theta)-C_k\|_2^2-R_k^2.
\label{eq:relative-score}
\end{equation}
When $s_{ik}<0$, the sample lies inside the hypersphere; otherwise, it lies on or outside the boundary. The IMHD-AD objective is
\begin{equation}
\begin{aligned}
\min_{\theta,\{C_k,R_k\}_{k=1}^{K}}
\mathcal{L}_{\mathrm{IMHD\text{-}AD}}
=
\sum_{k=1}^{K}
\Bigg[
&
R_k^2
+
\frac{1}{\nu n_k^{+}}
\sum_{i\in\mathcal{I}_{k}^{+}}
[s_{ik}]_{+}
+
\frac{1}{\mu n_k^{-}}
\sum_{i\in\mathcal{I}_{k}^{-}}
[-s_{ik}]_{+}
\Bigg] \\
&+
\frac{\lambda}{2}
\sum_{\ell=1}^{L}
\|\Theta_{\ell}\|_F^2,
\end{aligned}
\label{eq:joint-objective}
\end{equation}
where $[a]_{+}=\max(0,a)$ and $\Theta_{\ell}$ denotes the weight matrix of the $\ell$th feature-network layer. The first term controls the hypersphere volume, the second penalizes target samples outside their corresponding hypersphere, the third penalizes non-target samples inside that hypersphere, and the final term regularizes the network weights. The hyperparameters $\nu$ and $\mu$ control the penalties for target samples outside the sphere and non-target samples inside it, respectively.

\subsection{Construction and optimization of the interpretable multi-hypersphere network}

To optimize the hypersphere parameters jointly with the feature network, we adopt the parameter-embedding principle of interpretable neural clustering \cite{JMLR:v23:19-497}. Expanding Eq.~\eqref{eq:relative-score} gives
\begin{equation}
\begin{aligned}
s_{ik}
=
\|\phi(x_i;\theta)\|_2^2
-2C_k^\top\phi(x_i;\theta)
+\|C_k\|_2^2-R_k^2.
\end{aligned}
\label{eq:distance-expansion}
\end{equation}
Define the weight and bias of the $k$th unit in the final network layer as
\begin{equation}
\mathbf{w}_k=-2C_k,
\qquad
b_k=\|C_k\|_2^2-R_k^2,
\label{eq:parameter-mapping}
\end{equation}
The corresponding output is
\begin{equation}
g_k(x_i)
=
\mathbf{w}_k^\top\phi(x_i;\theta)+b_k.
\label{eq:last-layer-output}
\end{equation}
The boundary score of a sample relative to the $k$th hypersphere can therefore be written as
\begin{equation}
s_{ik}
=
g_k(x_i)+\|\phi(x_i;\theta)\|_2^2.
\label{eq:network-score}
\end{equation}

\begin{figure}[htbp]
    \centering
    \includegraphics[width=\linewidth]{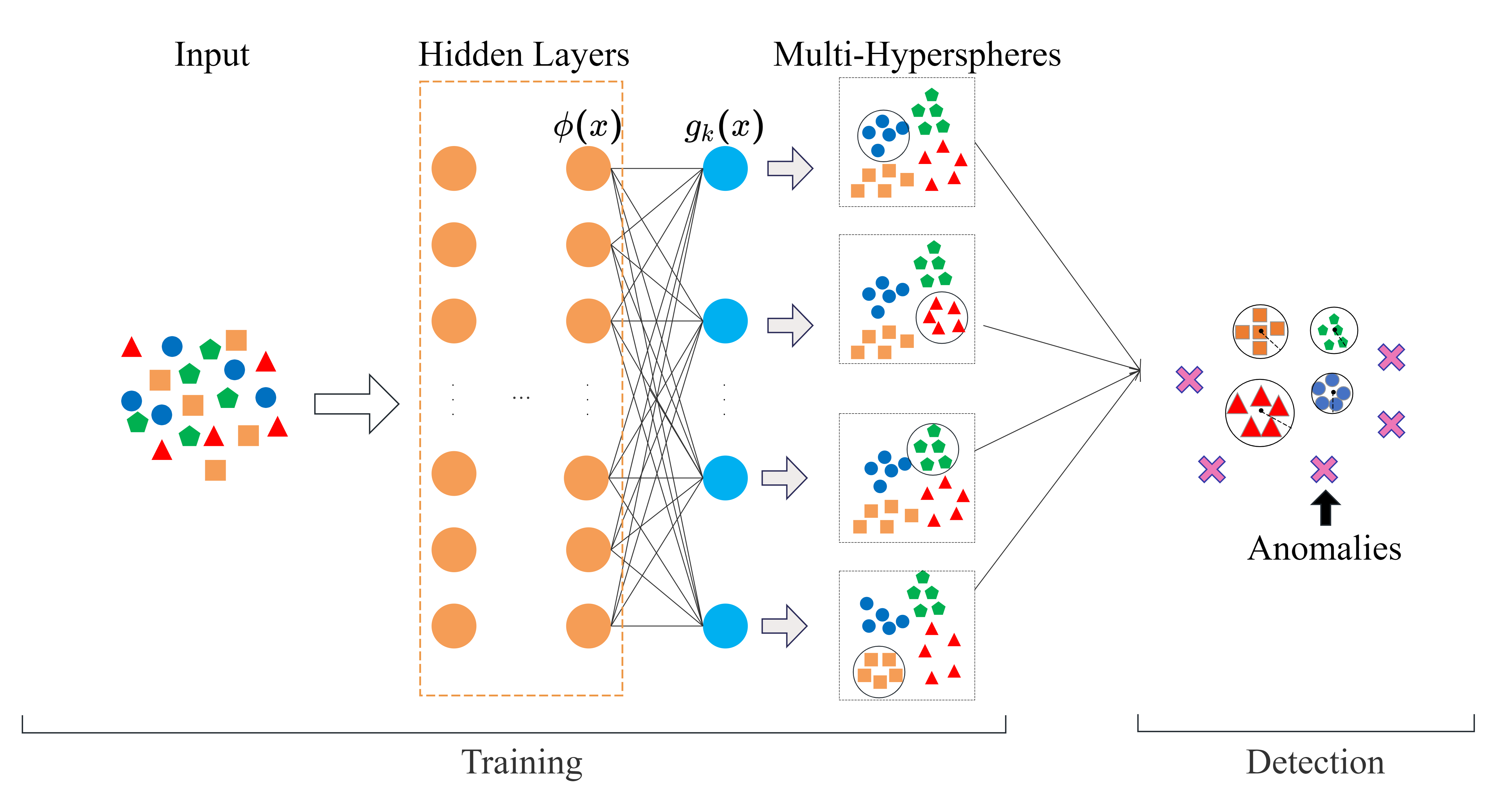}
    \caption{Architecture of the proposed interpretable multi-hypersphere deep anomaly-detection model.}
    \label{fig:imhd-ad-architecture}
\end{figure}

Fig.~\ref{fig:imhd-ad-architecture} illustrates the IMHD-AD architecture. The preceding layers learn the feature representation, while the final layer encodes the class-specific centers and radii. Equation~\eqref{eq:network-score} then measures the position of a sample relative to each hypersphere boundary.

Because $\mathbf{w}_k$ and $b_k$ are coupled, we impose the unit-norm constraint $\|C_k\|_2^2=1$ on each center. Combining this constraint with Eq.~\eqref{eq:parameter-mapping} yields
\begin{equation}
\|\mathbf{w}_k\|_2^2=4,
\qquad
C_k=-\frac{1}{2}\mathbf{w}_k,
\qquad
R_k^2=1-b_k.
\label{eq:recoverable-parameters}
\end{equation}
The additional constraint $b_k\leq1$ is required to ensure $R_k^2\geq0$.

Substituting Eq.~\eqref{eq:recoverable-parameters} into Eq.~\eqref{eq:joint-objective} gives the objective in terms of the network parameters:
\begin{equation}
\begin{aligned}
\min_{\theta,\{\mathbf{w}_k,b_k\}_{k=1}^{K}}
\widetilde{\mathcal{L}}_{\mathrm{IMHD\text{-}AD}}
=
\sum_{k=1}^{K}
\Bigg[
&
(1-b_k)
+
\frac{1}{\nu n_k^{+}}
\sum_{i\in\mathcal{I}_{k}^{+}}
[s_{ik}]_{+}
+
\frac{1}{\mu n_k^{-}}
\sum_{i\in\mathcal{I}_{k}^{-}}
[-s_{ik}]_{+}
\Bigg] \\
&+
\frac{\lambda}{2}
\sum_{\ell=1}^{L}
\|\Theta_{\ell}\|_F^2,
\end{aligned}
\label{eq:network-objective}
\end{equation}
subject to
\begin{equation}
\|\mathbf{w}_k\|_2^2=4,
\qquad
b_k\leq1,
\qquad
k=1,\ldots,K.
\label{eq:network-constraints}
\end{equation}

Introducing Lagrange multipliers $\alpha_k\in\mathbb{R}$ and $\beta_k\geq0$, the constraints are incorporated into the Lagrangian as
\begin{equation}
\mathcal{J}
=
\widetilde{\mathcal{L}}_{\mathrm{IMHD\text{-}AD}}
+
\sum_{k=1}^{K}
\left[
\alpha_k
\left(
\|\mathbf{w}_k\|_2^2-4
\right)
+
\beta_k(b_k-1)
\right].
\label{eq:lagrangian}
\end{equation}
Training corresponds to the following primal--dual problem:
\begin{equation}
\max_{\substack{\alpha_k\in\mathbb{R}\\\beta_k\geq0}}
\;
\min_{\theta,\{\mathbf{w}_k,b_k\}_{k=1}^{K}}
\mathcal{J}.
\label{eq:primal-dual}
\end{equation}
The network parameters $\theta$, $\mathbf{w}_k$, and $b_k$ are updated by gradient descent, while the Lagrange multipliers are updated by gradient ascent. The gradients with respect to the dual variables are
\begin{equation}
\frac{\partial\mathcal{J}}{\partial\alpha_k}
=
\|\mathbf{w}_k\|_2^2-4,
\qquad
\frac{\partial\mathcal{J}}{\partial\beta_k}
=
b_k-1.
\label{eq:dual-gradient}
\end{equation}
After each update, $\beta_k$ is projected onto the nonnegative real line to maintain $\beta_k\geq0$.

Table~\ref{tab:imhd-ad-algorithm} summarizes the IMHD-AD training procedure.

\begin{table}[htbp]
    \centering
    \caption{Training procedure for IMHD-AD.}
    \label{tab:imhd-ad-algorithm}
    \small
    \setlength{\tabcolsep}{3pt}
    \renewcommand{\arraystretch}{1.2}
    \begin{tabular}{@{}r p{0.82\linewidth}@{}}
        \toprule
        \textit{Input}
        &
        Training set $\mathcal{D}$; number of classes $K$; hyperparameters $\nu$, $\mu$, and $\lambda$; learning rates $\eta_{\mathrm{p}}$ and $\eta_{\mathrm{d}}$; and number of epochs $T$.\\
        \textit{Output}
        &
        Network parameters $\theta$ and $\{\mathbf{w}_k,b_k\}_{k=1}^{K}$; hypersphere parameters $\{C_k,R_k\}_{k=1}^{K}$.\\
        \midrule
        1 & Initialize the network parameters and Lagrange multipliers $\alpha_k$ and $\beta_k$.\\
        2 & \textbf{for} epoch $=1$ \textbf{to} $T$ \textbf{do}\\
        3 & \quad Sample a mini-batch and compute $z_i=\phi(x_i;\theta)$.\\
        4 & \quad Compute $g_k(x_i)=\mathbf{w}_k^\top z_i+b_k$.\\
        5 & \quad Compute $s_{ik}=g_k(x_i)+\|z_i\|_2^2$.\\
        6 & \quad Evaluate $\mathcal{J}$ using Eq.~\eqref{eq:lagrangian}.\\
        7 & \quad Update $\theta$, $\mathbf{w}_k$, and $b_k$ by gradient descent.\\
        8 & \quad Update $\alpha_k$ and $\beta_k$ by gradient ascent and project $\beta_k$ onto the nonnegative real line.\\
        9 & \textbf{end for}\\
        10 & Recover the hypersphere parameters using $C_k=-\mathbf{w}_k/2$ and $R_k=\sqrt{\max(0,1-b_k)}$.\\
        \bottomrule
    \end{tabular}
\end{table}

\subsection{Anomaly decision and geometric interpretation}

For a test sample $x$, the model computes its score relative to the $k$th hypersphere boundary as
\begin{equation}
s_k(x)
=
\|\phi(x;\theta)-C_k\|_2^2-R_k^2
=
g_k(x)+\|\phi(x;\theta)\|_2^2.
\label{eq:test-score}
\end{equation}
The overall anomaly score is defined as
\begin{equation}
f(x)
=
\min_{k\in\{1,\ldots,K\}}s_k(x),
\label{eq:anomaly-score}
\end{equation}
and the class with the minimum boundary score is
\begin{equation}
k^{*}(x)
=
\arg\min_{k\in\{1,\ldots,K\}}s_k(x).
\label{eq:class-prediction}
\end{equation}

If $f(x)<0$, the sample lies inside at least one known-class hypersphere and is assigned to class $k^{*}(x)$. If $f(x)\geq0$, it lies outside every hypersphere and is identified as anomalous. When a sample lies inside multiple hyperspheres, it is assigned to the class with the smallest boundary score.

This decision process also provides the geometric interpretation of the model. The center $C_k$ serves as the prototype of class $k$ in the feature space, $R_k$ specifies its acceptance range, and $s_k(x)$ indicates the position of a sample relative to the corresponding boundary. Comparing $\{s_k(x)\}_{k=1}^{K}$ reveals which class boundary is closest, whether the sample enters the associated hypersphere, and why it is rejected as anomalous. Because the explanation is derived directly from the quantities used for prediction, no additional post-hoc explainer is required.

\section{Experiments}
This section evaluates IMHD-AD on multi-hypersphere anomaly-detection tasks constructed from MNIST, Fashion-MNIST, and CIFAR-10. A two-dimensional synthetic dataset is additionally used to examine the geometry of the learned representation.

\subsection{Datasets}

We conduct benchmark experiments on three public image datasets: MNIST, Fashion-MNIST, and CIFAR-10. Their basic statistics are summarized in Table~\ref{tab:dataset-summary}.

\begin{table}[htbp]
    \centering
    \small
    \setlength{\tabcolsep}{4pt}
    \caption{Summary of the public benchmark datasets.}
    \label{tab:dataset-summary}
    \resizebox{\linewidth}{!}{%
    \begin{tabular}{cccc}
        \toprule
        \textbf{Dataset}
        & \textbf{Classes}
        & \textbf{Number of samples}
        & \textbf{Image size}\\
        \midrule
        MNIST
        & 10
        & 60,000 (train) + 10,000 (test)
        & $28\times28$ (grayscale)\\
        Fashion-MNIST
        & 10
        & 60,000 (train) + 10,000 (test)
        & $28\times28$ (grayscale)\\
        CIFAR-10
        & 10
        & 50,000 (train) + 10,000 (test)
        & $32\times32$ (RGB)\\
        \bottomrule
    \end{tabular}%
    }
\end{table}

\textbf{MNIST.} MNIST consists of ten classes of handwritten digits from 0 to 9. Each sample is a $28\times28$ single-channel grayscale image. The dataset contains 60,000 training images and 10,000 test images \cite{lecun1998gradient}.

\textbf{Fashion-MNIST.} Fashion-MNIST contains ten clothing classes: T-shirt/top, trouser, pullover, dress, coat, sandal, shirt, sneaker, bag, and ankle boot. Its training and test-set sizes and image dimensions match those of MNIST, but its classes exhibit more complex variations in texture and shape \cite{xiao2017fashion}.

\textbf{CIFAR-10.} CIFAR-10 contains 60,000 $32\times32$ color images from ten natural-image classes: airplane, automobile, bird, cat, deer, dog, frog, horse, ship, and truck. Of these images, 50,000 are used for training and 10,000 for testing \cite{krizhevsky2009learning}.

For each public dataset, we construct multi-hypersphere anomaly-detection tasks by varying the numbers of known normal and unseen anomaly classes. For any class-specific hypersphere, samples from that class are treated as target samples and samples from the other known normal classes as non-target samples.

\subsection{Compared methods and implementation details}

We compare IMHD-AD with four representative multi-class or multi-hypersphere anomaly-detection methods.

\begin{itemize}
    \item \textbf{MC-SVDD} \cite{mu2009multiclass} extends SVDD to multi-class data description and constructs a decision boundary for each class. The parameter $\nu$ is selected from $\{0.1,0.2,\ldots,0.9\}$, the penalty factor $C$ from $\{2^{-5},2^{-3},\ldots,2^{7}\}$, and the default setting is used for the kernel parameter $\gamma$.
    
    \item \textbf{DeepSVDD} \cite{ruff2018deep} learns a nonlinear representation with a deep neural network and constructs a compact hypersphere in the feature space. The parameter $\nu$ is selected from $\{0.01,0.05,0.1,\ldots,0.9\}$.
  
    \item \textbf{DMSVDD} \cite{ghafoori2020deep} extends DeepSVDD to multi-hypersphere data description and models complex distributions by jointly optimizing a deep network and multiple hyperspheres. The parameter $\nu$ is selected from $\{0.1,0.2,\ldots,0.9\}$.

    \item \textbf{DeepMAD} \cite{singh2022multi} treats the other known classes as pseudo-anomalies relative to the current class and learns discriminative class representations using a distance margin. The parameter $\lambda$ is selected from $\{0.001,0.01,0.1,1,10\}$.

    \item \textbf{IMHD-AD (proposed)} constructs an independent hypersphere for each class in a shared feature space, constrains target samples to lie inside the corresponding sphere and non-target samples to lie outside it, and jointly optimizes all centers and radii.
\end{itemize}

The open-set experiments compare MC-SVDD, DeepSVDD, DMSVDD, DeepMAD, and IMHD-AD under identical data splits and evaluation protocols. Hyperparameters are selected on validation data; the test set is used neither for model training nor for parameter selection.

To accommodate differences in image complexity, we use convolutional feature extractors of different capacities, as detailed in Table~\ref{tab:network-settings}. All convolutional layers use $3\times3$ kernels with a stride of 1 and padding of 1. Each max-pooling layer uses a $2\times2$ kernel with a stride of 2. ReLU activations are applied after convolutional and fully connected layers \cite{nair2010rectified}.

\begin{table}[htbp]
    \centering
    \small
    \setlength{\tabcolsep}{4pt}
    \caption{Network configurations for the public image datasets.}
    \label{tab:network-settings}
    \resizebox{\linewidth}{!}{%
    \begin{tabular}{ccccc}
        \toprule
        \textbf{Dataset}
        & \textbf{Conv. layers}
        & \textbf{Conv. channels}
        & \textbf{Embedding dim.}
        & \textbf{Dropout}\\
        \midrule
        MNIST
        & 2
        & 6, 16
        & 64
        & 0.2\\
        Fashion-MNIST
        & 2
        & 8, 24
        & 96
        & 0.3\\
        CIFAR-10
        & 3
        & 32, 64, 128
        & 256
        & 0.5\\
        \bottomrule
    \end{tabular}%
    }
\end{table}

The feature extractors for MNIST and Fashion-MNIST contain two convolutional and two pooling layers. For CIFAR-10, we use three convolutional and three pooling layers and apply batch normalization after each convolutional layer \cite{ioffe2015batchnormalization}. Dropout is used in the fully connected layers to mitigate overfitting \cite{srivastava2014dropout}. MNIST images are normalized to $[0,1]$. Fashion-MNIST additionally uses random horizontal flipping, while CIFAR-10 images are standardized channel-wise and augmented by random cropping and horizontal flipping.

IMHD-AD is trained using the Adam optimizer \cite{kingma2015adam}. Its principal experimental settings are listed in Table~\ref{tab:hyperparameters}.

\begin{table}[htbp]
    \centering
    \small
    \caption{Experimental hyperparameters for IMHD-AD.}
    \label{tab:hyperparameters}
    \begin{tabular}{cll}
        \toprule
        \textbf{Category} & \textbf{Parameter} & \textbf{Setting}\\
        \midrule
        \multirow{7}{*}{Training}
        & Random seed & 42\\
        & Optimizer & Adam\\
        & Initial learning rate & $3.0\times10^{-4}$\\
        & Weight decay & $0.5\times10^{-6}$\\
        & Maximum epochs & 200\\
        & Batch size & 200\\
        \midrule
        \multirow{3}{*}{Hypersphere}
        & Learning-rate step & 50 epochs\\
        & Parameter $\nu$ & $\{0.1,0.3,0.5,0.7,0.9\}$\\
        & Parameter $\mu$ & $\{0.1,0.3,0.5,0.7,0.9\}$\\
        \bottomrule
    \end{tabular}
\end{table}

We consider two open-set protocols on the public datasets: two known normal classes with eight unseen anomaly classes, and three known normal classes with seven unseen anomaly classes.

\subsection{Results and analysis}

The area under the receiver operating characteristic curve (AUC) is used as the primary evaluation metric. AUC measures the ability of a model to distinguish target from non-target samples across decision thresholds; a higher value indicates better overall discrimination \cite{fawcett2006introduction}. For each multi-class task, we compute a one-vs-rest AUC for every known normal class and report their arithmetic mean.

\subsubsection{Multi-hypersphere anomaly-detection benchmarks}

To evaluate the multi-hypersphere modeling capability of IMHD-AD, we construct open-set tasks on MNIST, Fashion-MNIST, and CIFAR-10 with different numbers of known normal and unseen anomaly classes. The model learns an independent hypersphere for each known normal class and uses samples from the other known classes to encourage separation among the hyperspheres.

In the first protocol, two of the ten classes in each dataset are selected as known normal classes and the remaining eight serve as unseen anomaly classes. Table~\ref{tab:two-normal-aucs} reports detailed results for five randomly selected class combinations. All values are AUC scores (\%), and the best result in each row is shown in bold.

\begin{table}[htbp]
\caption{AUC comparison on MNIST, CIFAR-10, and Fashion-MNIST with two known normal classes and eight unseen anomaly classes.}
\label{tab:two-normal-aucs}
\centering
\small
\setlength{\aboverulesep}{0.1ex}
\setlength{\belowrulesep}{0.1ex}
\renewcommand{\arraystretch}{1.2}
\setlength{\tabcolsep}{6pt}
\resizebox{\textwidth}{!}{%
\begin{tabular}{llccccc}
\toprule
\textbf{Dataset} & \textbf{Known-class combination} & \textbf{MC-SVDD}
& \textbf{DeepSVDD} & \textbf{DMSVDD} & \textbf{DeepMAD} & \textbf{IMHD-AD}\\
\midrule
\multirow{5}{*}{MNIST}
& (0, 2) & 91.23 & 86.14 & \textbf{97.64} & 94.07 & 96.77\\
& (0, 6) & 94.96 & 96.52 & 88.75 & 96.56 & \textbf{96.99}\\
& (0, 7) & 93.71 & 87.68 & \textbf{97.15} & 96.85 & 95.40\\
& (0, 9) & 91.83 & 91.85 & 87.96 & 91.61 & \textbf{97.25}\\
& (4, 9) & 92.85 & 95.23 & 83.89 & 83.70 & \textbf{97.50}\\
\midrule
\multirow{5}{*}{CIFAR-10}
& (0, 2) & 50.19 & 47.32 & 71.63 & 57.12 & \textbf{90.28}\\
& (0, 6) & 53.46 & 46.53 & 73.05 & 76.10 & \textbf{90.12}\\
& (0, 7) & 57.43 & 51.32 & 63.24 & 70.82 & \textbf{89.29}\\
& (0, 9) & 61.94 & 61.31 & 75.66 & 75.42 & \textbf{91.41}\\
& (4, 9) & 56.21 & 54.02 & 65.30 & 70.74 & \textbf{90.38}\\
\midrule
\multirow{5}{*}{Fashion-MNIST}
& (0, 2) & 70.59 & 80.33 & 81.66 & 83.30 & \textbf{96.73}\\
& (0, 6) & 79.47 & 81.66 & 82.58 & 67.46 & \textbf{95.10}\\
& (0, 7) & 83.07 & 86.55 & 76.82 & 83.51 & \textbf{97.69}\\
& (0, 9) & 81.59 & 86.67 & 72.07 & 79.11 & \textbf{97.32}\\
& (4, 9) & 83.06 & 87.56 & 89.23 & 87.90 & \textbf{96.85}\\
\bottomrule
\end{tabular}%
}
\end{table}

As shown in Table~\ref{tab:two-normal-aucs}, IMHD-AD achieves the highest AUC in 13 of the 15 comparisons. It ranks first for three of the five MNIST combinations and for all combinations on CIFAR-10 and Fashion-MNIST.

The second protocol increases the number of known normal classes to three and treats the remaining seven classes as unseen anomalies. Results for five randomly selected combinations are reported in Table~\ref{tab:three-normal-aucs}.

\begin{table}[htbp]
\caption{AUC comparison on MNIST, CIFAR-10, and Fashion-MNIST with three known normal classes and seven unseen anomaly classes.}
\label{tab:three-normal-aucs}
\centering
\small
\setlength{\aboverulesep}{0.1ex}
\setlength{\belowrulesep}{0.1ex}
\renewcommand{\arraystretch}{1.2}
\setlength{\tabcolsep}{6pt}
\resizebox{\textwidth}{!}{%
\begin{tabular}{llccccc}
\toprule
\textbf{Dataset} & \textbf{Known-class combination} & \textbf{MC-SVDD}
& \textbf{DeepSVDD} & \textbf{DMSVDD} & \textbf{DeepMAD} & \textbf{IMHD-AD}\\
\midrule
\multirow{5}{*}{MNIST}
& (2, 6, 9) & 78.38 & 78.66 & 81.32 & 96.13 & \textbf{97.45}\\
& (3, 6, 7) & 73.60 & 76.76 & 78.03 & 96.60 & \textbf{96.68}\\
& (0, 8, 9) & 89.04 & 87.55 & 92.09 & 92.14 & \textbf{96.51}\\
& (0, 6, 8) & 89.43 & 88.09 & 86.44 & 87.29 & \textbf{93.92}\\
& (0, 5, 8) & 85.23 & 84.46 & 89.22 & 94.82 & \textbf{97.61}\\
\midrule
\multirow{5}{*}{CIFAR-10}
& (2, 6, 9) & 51.02 & 46.68 & 69.48 & 78.92 & \textbf{88.02}\\
& (3, 6, 7) & 53.53 & 50.41 & 73.03 & 80.93 & \textbf{88.26}\\
& (0, 8, 9) & 67.13 & 66.98 & 73.36 & 82.72 & \textbf{88.78}\\
& (0, 6, 8) & 62.43 & 57.61 & 76.56 & 80.38 & \textbf{90.79}\\
& (0, 5, 8) & 58.98 & 64.11 & 62.74 & 71.19 & \textbf{90.39}\\
\midrule
\multirow{5}{*}{Fashion-MNIST}
& (2, 6, 9) & 79.79 & 72.13 & 80.83 & 72.37 & \textbf{96.85}\\
& (3, 6, 7) & 72.63 & 77.72 & 83.23 & 70.62 & \textbf{97.17}\\
& (0, 8, 9) & 67.76 & 78.93 & 86.87 & 83.07 & \textbf{95.72}\\
& (0, 6, 8) & 71.11 & 75.73 & 75.89 & 60.47 & \textbf{97.07}\\
& (0, 5, 8) & 54.29 & 60.93 & 79.41 & 74.46 & \textbf{96.58}\\
\bottomrule
\end{tabular}%
}
\end{table}

As shown in Table~\ref{tab:three-normal-aucs}, IMHD-AD achieves the highest AUC in all 15 comparisons, indicating that it can learn multiple class-specific hyperspheres collaboratively.

\subsection{Visualization analysis}

To investigate the effect of network complexity on multi-hypersphere anomaly detection, we construct a two-dimensional synthetic dataset. The training set contains four normal classes generated from Gaussian, uniform, Laplace, and mixture distributions, respectively, with each class located around a different center in the two-dimensional space. The test set retains these four normal classes and additionally includes an arc-shaped anomaly class that is absent during training. Although the center of this anomaly class lies close to one normal class, its overall distribution has a different shape. It therefore represents an unknown anomaly that is locally close to a known normal class. We keep the data, random seed, and training hyperparameters fixed and vary only the architecture of the feature network. Detection performance is evaluated using anomaly-detection AUC.

\begin{figure}[htbp]
    \centering
    \begin{minipage}[t]{0.49\linewidth}
        \centering
        \includegraphics[width=\linewidth]{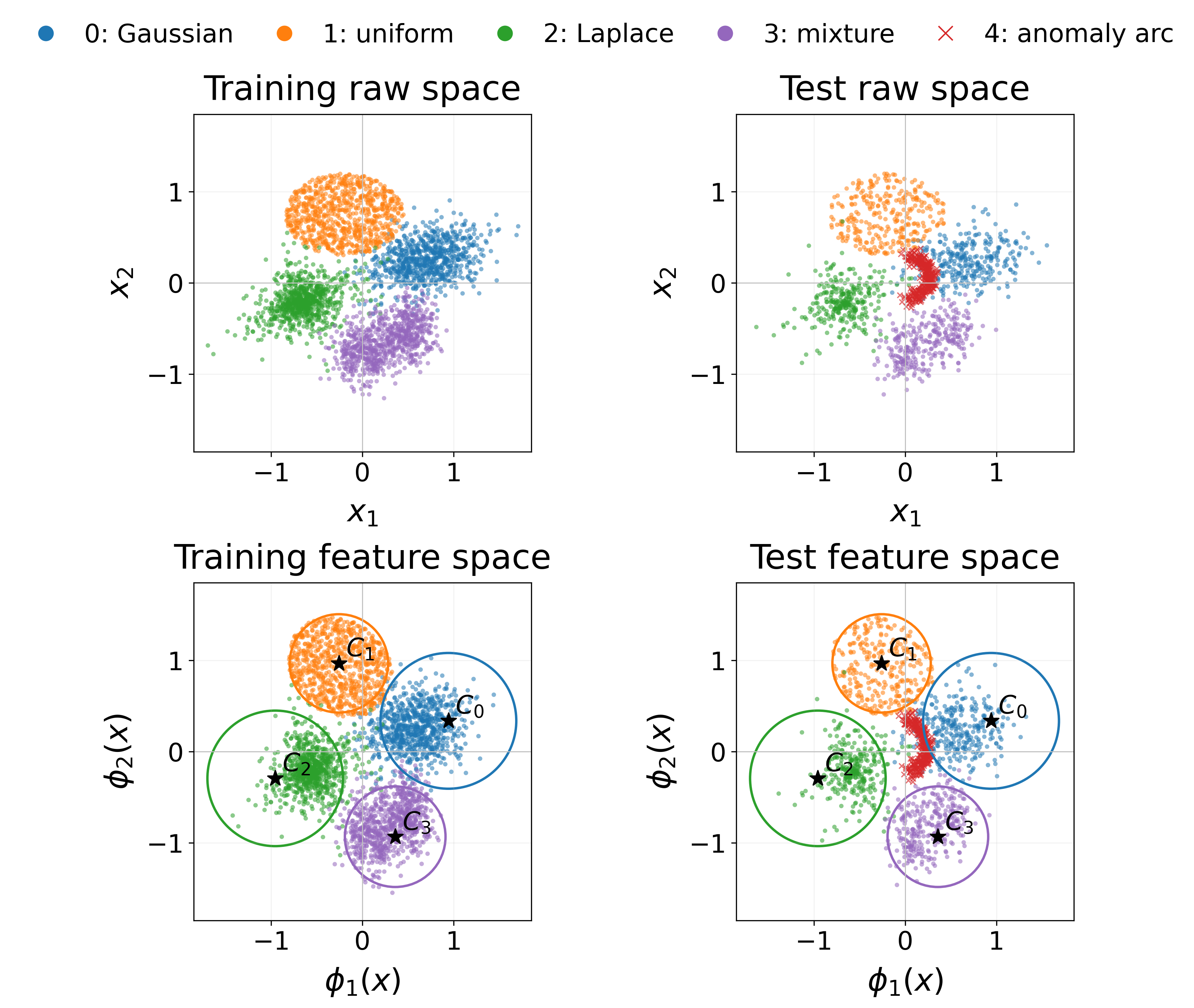}
        \vspace{0.2em}
        \small
        \textbf{(a) Linear network: $2\times2$}
    \end{minipage}
    \hfill
    \begin{minipage}[t]{0.49\linewidth}
        \centering
        \includegraphics[width=\linewidth]{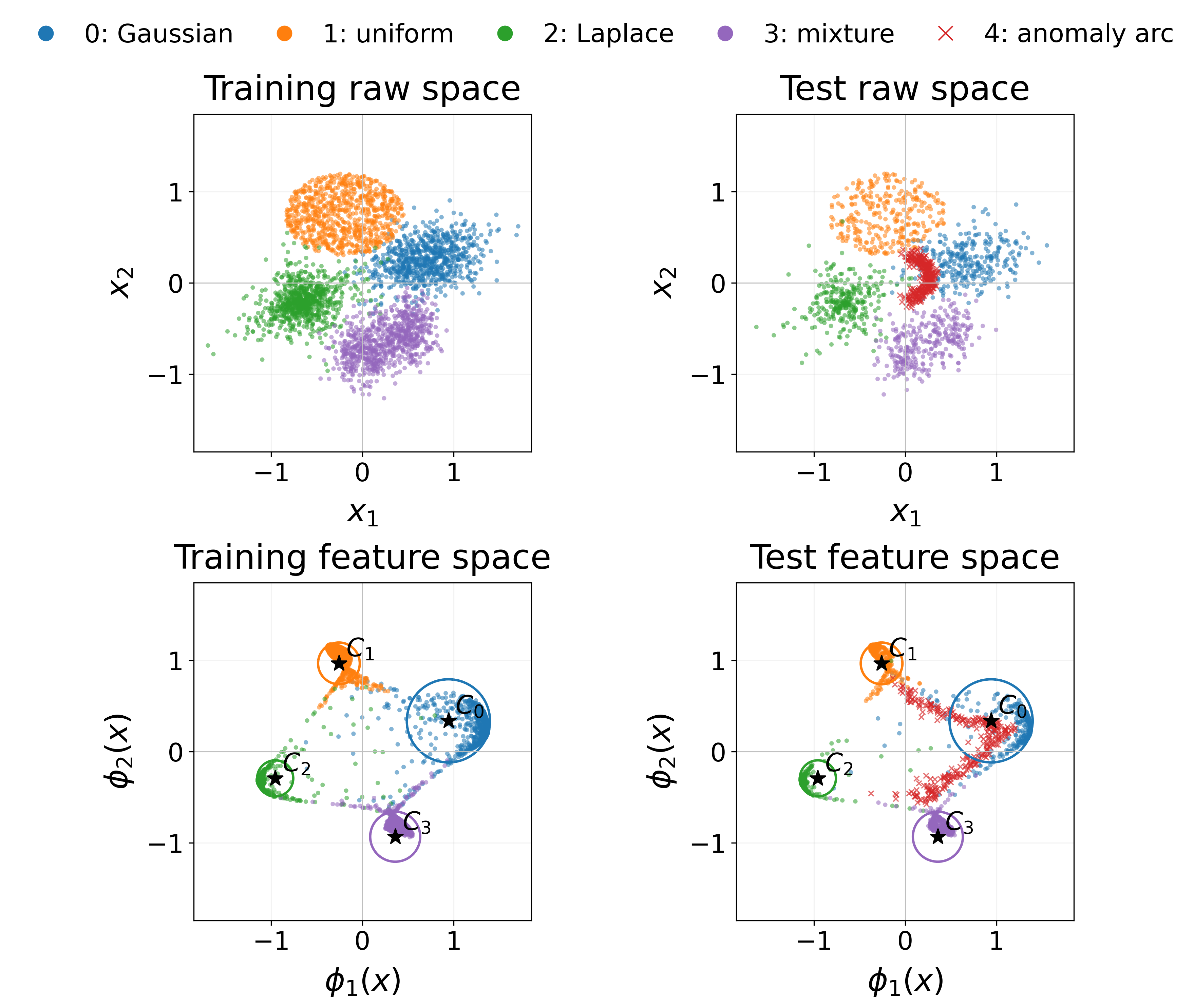}
        \vspace{0.2em}
        \small
        \textbf{(b) Deep network: $2\times32\times32\times16\times2$}
    \end{minipage}
    \caption{Feature-space visualizations of the synthetic data under different network complexities.}
    \label{fig:synthetic-complexity}
\end{figure}

Fig.~\ref{fig:synthetic-complexity} visualizes the feature-space geometry learned by a linear network and a deep network.

As shown in Fig.~\ref{fig:synthetic-complexity}, the linear network largely preserves the geometry of the original data, maintaining clear spatial relationships among the normal classes and effective separation between normal samples and unknown anomalies. The deep network compresses the normal classes more strongly and attains a normal-class classification accuracy of 0.9742. However, its anomaly-detection AUC decreases to 0.5780, while the average hypersphere radius contracts from 0.6443 to 0.2890.

Fig.~\ref{fig:synthetic-complexity}(b) also demonstrates the diagnostic value of the proposed interpretable network. In this two-dimensional setting, the learned features, hypersphere centers, and radii can be visualized directly without a post-hoc explainer. The visualization reveals that the deeper mapping folds unseen anomalies toward the normal hyperspheres while making these hyperspheres more compact. It therefore provides an intuitive explanation for the reduced detection performance on unseen classes at test time. This observation also highlights the need to balance the compactness of known normal classes against the separability of unknown anomalies.

\section{Conclusion}
We proposed Interpretable Multi-Hypersphere Deep Anomaly Detection (IMHD-AD) to address insufficient non-target constraints, ambiguous class boundaries, and limited decision transparency in multi-class anomaly detection. The method constrains target samples to lie inside their corresponding hyperspheres and non-target samples to remain outside, while jointly learning class-specific centers and radii. By mapping the hypersphere parameters to the weights and biases of the final network layer, IMHD-AD enables end-to-end optimization of the feature representation and multi-hypersphere boundaries. Class recognition and anomaly detection are based on the position of a sample relative to each boundary, allowing every decision to be explained directly in terms of the learned centers, radii, and boundary scores. Open-set experiments on MNIST, Fashion-MNIST, and CIFAR-10 show that IMHD-AD generally outperforms the compared methods. The synthetic visualization further demonstrates that model complexity must balance the compactness of known normal classes against the separability of unknown anomalies. The current method learns its acceptance boundaries exclusively from class-labeled normal data. Future work will investigate semi-supervised extensions that incorporate a small number of labeled anomaly samples during training to further refine the boundaries between normal and anomalous regions.

\bibliographystyle{elsarticle-num}
\bibliography{references_en}
\end{document}